# Toward Intelligent Biped-Humanoids Gaits Generation

Nizar ROKBANI , Boudour AMMAR CHERIF  and Adel M.  ALIMI
*nizar.rokbani@ieee.com , Boudour.ammar@gmail.com, adel.alimi@ieee.org*
*REGIM, Research Group on Intelligent Machines, National School of Engineering of Sfax,
University Of Sfax, Bp W, Soukra road, 3038 Sfax, Tunisia.*

**Abstract**

In this chapter we will highlight our experimental studies on natural human walking analysis and introduce a biologically inspired design for simple bipedal locomotion system of humanoid robots. Inspiration comes directly from human walking analysis and human muscles mechanism and control. A hybrid algorithm for walking gaits generation is then proposed as an innovative alternative to classically used kinematics and dynamic equations solving, the gaits include knee, ankle and hip trajectories. The proposed algorithm is an intelligent evolutionary based on particle swarm optimization paradigm. This proposal can be used for small size humanoid robots, with a knee an ankle and a hip and at least six Degrees of Freedom (DOF).

## 1. Introduction

In near future humanoid robots will be asked to "live" and collaborate with humans, they have generally a friendly design with a great resemblance with us. Biped humanoids robots are expected to have increasing field of exploitations, they are naturally adapted to human like commodities and have the advantage to fit more the human's environment than other kind of robots; humanoids can coordinate  tasks with humans workers (Arbulu & Balaguer, 2008) with human like comportments. Humanoid robotics issues include building machines with human like capabilities in motion control, gestures, and postures. Recently a lot of work is done in order to enhance their capacities in walking, fast walking and motion optimization (Xie et al, 2008) in a human like environment. The development of machine intelligence is crucial of the independence of these robots it is also crucial to the emergence of self learning attitudes. All these factors impulses the rapid growth of humanoid robotics.
Humanoid robots are legged robots with a limited number of two legs, such a limitation is important since it has direct impact on stability control policy. When two legs are used in a dynamic walking process, the sustention polygon is limited to a single footprint. In natural human walking process the double support phase is rarely used (Rokbani et al, 2008).

In legged robotics gaits analyses are commonly used to understand the locomotion system. Gaits analysis is a methodology introduced by biomechanics scientists to analyze and enhance motion dynamics of humans and animals. The human walking is the result of a long learning process that begins in the childhood life. Humans naturally adapt their walking-steps to their environments in order to ensure their safety and avoid falling down; they naturally optimize their energy and maintain their anthropomorphic stand up while switching between the support and transfer phases or between static and dynamic walking. The walking process takes into consideration all the humans activities, they can walk while eating, drinking or simply responding to phone call.

The "IZIMAN" is a research project of the REGIM laboratory, "Research group on intelligent Machines", with some ambitious challenges aiming to propose intelligent architectures that are biologically inspired in order to offer to humanoid robots a form of self tutoring (Rokbani et al, 2007).

In second section of this chapter we will highlight briefly the main issues of bipedal and humanoid locomotion robotics; in section III we present our gaits capture experiments, note that the gaits are extracted to be used as a comparison material, in section IV, we propose a hybrid method for human like gaits generation. The last section, V, is reserved to discussions and further works.

## 2. Humanoid locomotion robotics

### 2.1. Humanoid and biped robots evolution

A humanoid robot is commonly assumed to be a biped otherwise some wheeled humanoid robots are proposed by researchers (Berns et al, 1999). Humanoid robotics includes all the aspects of human like machines such as: walking, grasping, emotion or cognition, etc. Humanoid locomotion robotics is interested only in the issues concerning gaits generation, stability control, walking energy optimization. Recently an increasing interest to humanoid robotics helps the development of more humanoids; small size humanoids are actually proposed for entertainments; some models can be used as low cost research validation platforms. One of the earliest projects in humanoid locomotion is the Waseda university, Japan where Kato and his team build their walker robot, WL1, since 1966. By 1984 a humanoid dynamic walker is proposed and called WL10 RD, this prototype is a 12(DOF)[1], walker robot. Wabian is a real humanoid robot having 52 (DOF), including arms and head and able to walk at a speed of 0.21 meter/second. On the other hand, some sophisticated humanoids robots are developed by Japanese industrial companies such as Sony with their QRIO humanoid robot and Honda with its well known ASSIMO. One of the most enhanced humanoid projects is certainly the HRP2 Japanese projects; this robot is comparable to ASSIMO, the HRP2 is a Human size robot and shape, with low energy needs and it is proposed free from the back-bag, used as battery container in ASSIMO. It is important to distinguish between full size humanoids and smaller ones. Full size humanoids such as HRP2 or ASSIMO have a size which is comparable to humans. Robots like Nimbro (Behnke et al, 2007) or Qrio have sizes that not exceed 1m. Recently a French amazing humanoid robot, called Nao, was proposed while still under development. The Nao robot was the official platform for the IEEE Humanoid conference and Robocup competitions.

---

[1]　　　DOF : Degree Of Freedom

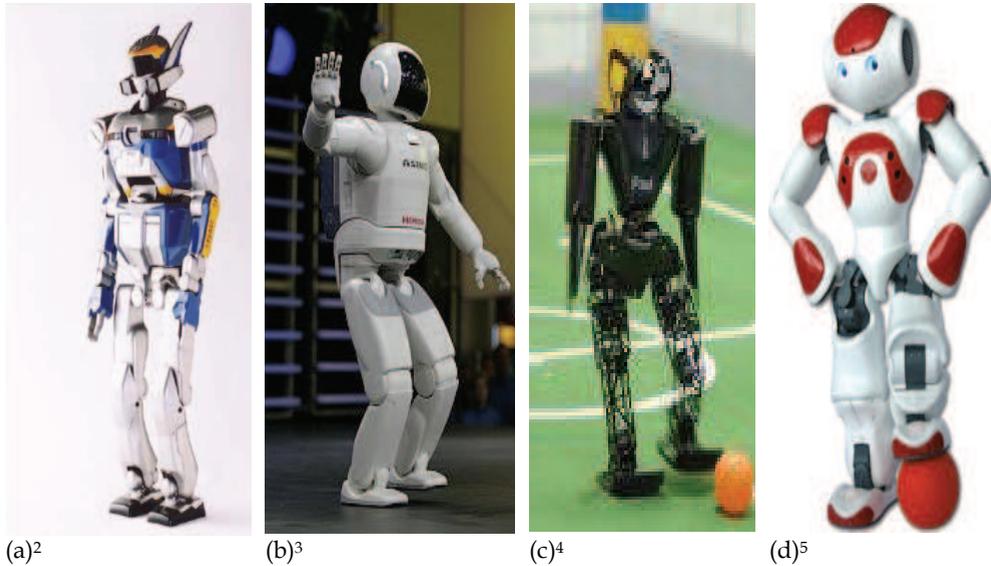

(a)[2]  (b)[3]  (c)[4]  (d)[5]

Fig. 1. Humanoid robots: (a) and (b) are human size robots, (c) and (d) are middle size robots. (a) HRP2, (b) ASSIMO, (c) Nimbro soccer robot, (d) NAO robot.

In Europe, some research projects have been developed in France, Belgium, Deutschland and Italy. French projects are essentially focused on the locomotion system. Bip and rabbit are the most famous projects; the "Bip" robot is an autonomous walker carrying his control system while the rabbit robot is a platform for dynamic walking analysis and experiments (Azevedo et al, 2004) (Chemori, 2005). In Belgium the Lucy robot is also a walker robot; its particularity is that it uses pneumatic muscles as actuators. Johnnie is a Deutsch humanoid founded by Technical University of Munich. Recently new proposals are made in order to generalize the ZMP[6] approach, new intuitive methods such us the (RTC)[7] (Goobon, 2008) allowing to maintain biped walker stability on irregular floor or stepping over an obstacle while walking (Sabourin et al, 2008).

---

[2]     Source : http://www.generalrobotix.com/
[3]     Source : http://www.androidworld.com
[4]     Source : www.informatik.uni-freiburg.de
[5]     Source : www.aldebaran-robotics.com
[6]     ZMP : Zero Moment Point
[7]     RTC : Relative Trajectory Control

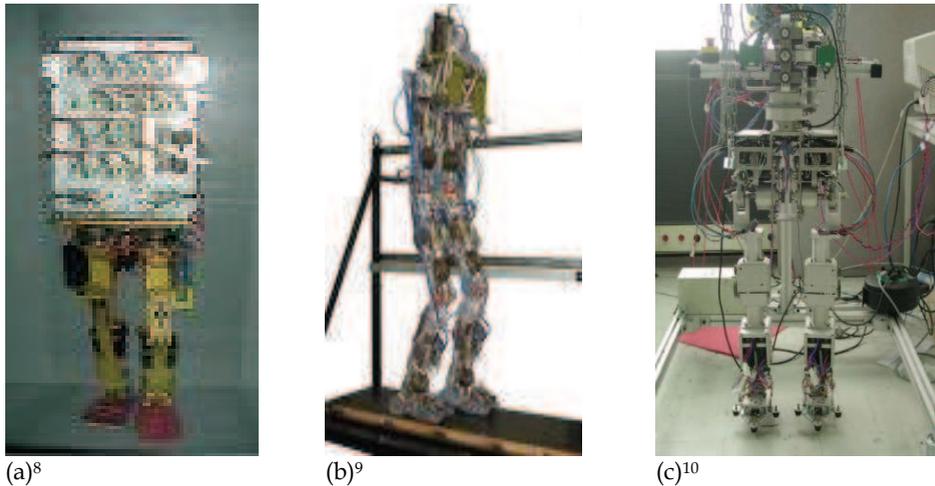

(a)[8]　　　　　　　　　　　　(b)[9]　　　　　　　　　　　　(c)[10]
Fig. 2. Humanoid locomotion systems or biped robots, (a) BIP robot, (b) Lucy robot, (C) Robian robot.

### 2.2. Description of the IZIMAN project

"IZIMAN" is a research project aiming to propose alternative intelligent solutions to humanoid robots. The IZIMAN architecture aims to propose a schema to help emergence of machine intelligence rather than to create an intelligent machine directly inspired from humans; the proposed architecture is a multi agent's one, composed by controls and decisions centers that have to collaborate to ensure the man-like tasks of the humanoid robot. A special focus is attributed to the learning processes; learning uses build in intelligent based on evolutionary, fuzzy or neuro-fuzzy algorithms. Human imitations are also considered as source of learning frames. Learning based on imitation needs efficient human detection and gestures tracking.

Assuming that the proposed control system is hybrid and integrates different types of controllers; a collaboration schema should be defined to avoid collateral risk of antagonist command signals. The collaboration schemas establish the hierarchy between the controllers and solve the multiplicity command problematic (Rokbani et al, 2007).

For the gaits generation of the biped locomotion system, we opted for a multi-swarms design, each swarm controls a joint. A collaboration schema insures the coherence on the generated joints trajectories called also gaits. This architecture will be detailed in section IV.

### 3. Human walking analysis

Human walking is a complex process that uses a 29 (DOF) locomotion system coupled with different neural centers including brain and spinal-reflex centers. More than 48 muscles are used to coordinate foot motions and synchronize them with the upper body constrains. The joints of each foot perform elementary movements essentially flexion and extension; that includes the hip, the knee and the ankle. The walking is a result from the coordination of these simple movements. It is important to note that upper body is connected to the locomotion system trough the basin; the upper body includes a critical point assumed to be the center of mass, COM, and playing a key role in balance, stability and walking dynamics.

---

[8]　　　　Source : http://www-lms.univ-poitiers.fr/article182.html
[9]　　　　Source : http://lucy.vub.ac.be
[10]　　　Source : http://www2.cnrs.fr/presse/communique/410.htm

### 3.1. Description of human locomotion system

From an anatomic point of view, the lower trunk of humans is composed by a pelvic bone connecting two legs. A leg is composed by two vertical segments: the femur and the tibia. The femur is attached to the pelvic bone by the hip joint and to the tibia by the knee joint (Wagner & Carlier, 2002). Biomechanical description of the human like lower trunk shows that the ankle performs two rotations, the knee performs two rotations and the hip three rotations. Most biped construction use limited degrees of freedom compared to human locomotion system. Such a limitation reduces the dynamic of the walking but makes easier the stability control. Human motions result from flexions and extensions; these elementary actions are produced by muscles. Assuming the effectiveness of the muscles, pneumatic muscles proposed and had been used in biped robots such as Lucy biped. In human locomotion system flexion and extension of muscles produces rotations on the human joints. In a humanoid locomotion system these rotations depend on the skeleton structure and on the actuators specification; but whatever are the nature of the build skeleton and the used actuators most bipeds respect globally the human joints angular elongations limits.

### 3.2. Description of walking gaits

The human body is represented by segments and joints; this simplification is commonly used in biomechanics; the outline of a leg can be represented by seven anatomical markers for three joints (Winter, 1990), a segment is obtained by joining two consecutive markers. Considering the complexity of human walk it is common to use a limited number of states to represent a walking cycle; only key steps are detailed and assumed to be representative of overall a cycle. Wagner uses sixteen states (Wagner & Carlier, 2002) while Winter uses thirteen (Winter, 1990). In robotics the used steps are limited compared to biomechanics descriptions (azevedo & Heliot, 2005). This representation simplifies the computing of angular limits of the joints, from a gait analysis it becomes possible to establish an estimation of these limits.

To collect human gaits, we have used the classical marking methodology where participants are young humans with ages ranging from 20 to 24 years, the participants are students at the High Institute of Sport and Physical Education of Sfax, Tunisia. We have installed a walking scene of about 8 meters long, 2 meters large and 4 m height. The scene walls were covered by black, to simplify marker extraction. Experiments were conducted under supervision of biomechanics' walking specialists. Six participants are asked to work normally from a start point to the end of the scene, avoiding accelerations and decelerations. Two cameras were placed respectively at the left of the walkers and in front of them. We had market the hip, the knee, the ankle and the foot of each walker using white circular stickers with a diameter of 3 cm (centimeter), see figure 3. The walking cycles were captured using two numeric cameras placed respectively at 5m40 from the lateral plan and at 8m 25 from the frontal plan. The scene, a lateral walker and a frontal walker can be seen in respectively figure 3 (a), (b) and (c).

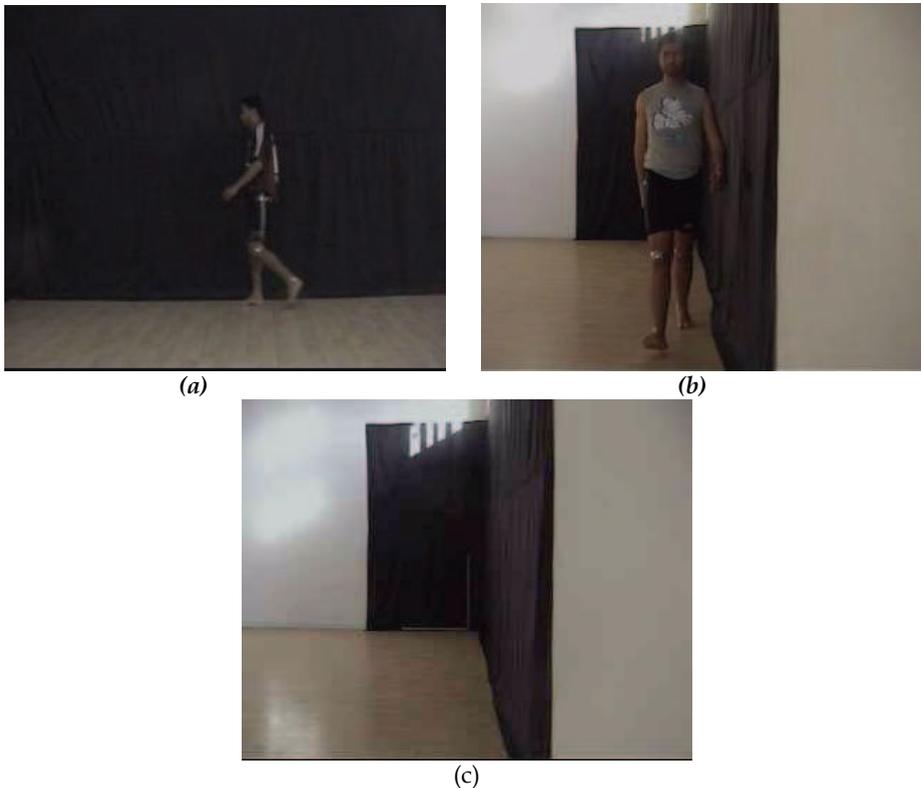

Fig. 3. human gaits capture, (a) human walking on lateral plan, (b) human walking in frontal plan, (c) capture scene.

## 4. Hybrid approach to humanoid like gaits generation

### 4.1 Walker initialization

Since we are concerned by human inspired locomotion system, we used human anthropometry to establish an approximation for the humanoid body segments dimensions. In human anthropometry theses dimensions depend on sex, body build, race origin, etc. In our work, we are using an average length expressed in (Winter, 1990) and detailed by formulas (1) to (5). A humanoid segment-based representation is then built as it can be seen in figure 4(c). It is important to indicate that in this work the footprint is added just in order to simplify the static stability condition; in real applications a footprint introduces the problem to floor reaction.

$$fl = 0.152 * h \tag{1}$$

$$fb = 0.055 * h \tag{2}$$

$$tl = 0.246 * h \tag{3}$$

$$leg\_l = 0.53 * h \tag{4}$$

$$Int\_h = 0.191 * h \tag{5}$$

Where *fl, fb* represent respectively the foot length and the foot breadth, *tl* is the length of the segment knee to ankle; the *leg_l* term represents the leg length and **Int_h** is the inter-hip distance.

After walker skeleton initialization we introduce a new proposal to achieve gaits generation; instead of using classical kinematics modeling we introduce a swarm based search policy. A swarm of particles is placed in the corresponds joints of the walker, the swarm has to find a suitable new position for the joint assuming both forward stepping and avoiding the robot fall down.

In particle swarm optimization, as founded by Kennedy and Eberhart (Kennedy & Eberhart, 1995) (Dreo et al, 2003), the swarm is composed by a set of particles; each particle has its position, its velocity and its fitness function. Initially the particles are randomly initialized round a joint position; a search space is then allocated to the swarm. Particles have to find within the search space the new position, all the swarm will consider its actual position and the position of the best particle. A fitness function is used to evaluate the performance of each particle and to select the best one. The original formulation of PSO can be expressed through the following formulas:

$$v_{i(t)} = v_{i(t-1)} + c_1 * r_1 * [p_{lbest} - x_{i(t-1)}] + c_2 * r_2 * [p_{gbest} - x_{i(t-1)}] \tag{6}$$

$$x_{i(t)} = x_{i(t-1)} + v_{i(t)} \tag{7}$$

Where, $x_{i(t)}$ is the current position of the particle (i), $v_{i(t)}$ is the particle velocity;

*c1* represents the moderation of particle personal contribution;

*c2* represents the moderation of social contribution; (r1) and (r2) are random numbers within the interval [0.0, 1]. Note that $p_{lbest}$ represents the best particle in the particle (i) neighborhood; $p_{gbest}$ is the particle getting the best fitness function of the swarm. (t) is the current time step and (t-1) is the last time step.

The swarms are grouped in right and left foot sub-swarms. Since in walking cycle we have a support foot and an oscillating one the foot sub-swarms will moves alternatively; when a foot is assumed to be the support one, all its particles are in hunt mode. Collaboration is also very important, a forward valid step depends on the joints positions of the overall skeleton. To satisfy those constraints some collaboration connections are established between the joints swarms; the connections appear in figure 4(b). Really a connection means that the best position of a joint is communicated to the upper joint and the virtual COM in order to check and validate the stability of the forthcoming joints displacements.

### 4.2 Swarms processing details

The proposed architecture is composed by two sub-swarms; the sub-swarm(0), see figure 4 (b), represents the left foot and is composed by three main particles *P0.1*, *P0.2* and *P0.3* respectively the hip, the knee and the ankle. The same organization is used for the right feet which is represented by the sub-swarm(1). These particles represent valid skeleton positions and are not supposed to move frequently, they are memories particles. Around each one of them, there is a limited set of search particles working in search space prospecting. A walking step includes seven coordinates representing the articulations and the body mass. To ensure the stability of the system we assign a memory-particle to each articulation. This particle will be moved only if its next position ensures a forward walking-step and the static stable attitude of the walker.

If we consider the particle *p0.1*, see figure 4 (b), this particle is a memory and will not be authorized to move since its new position is not better than its actual one and since the new position do not insure stability. The search particles related to *p0.1* are initialized and iterates looking for the best position in their allocated search space. The local best position is elected according to both the local search strategy and with respect to global stability strategy of the walker. Global stability depends on whether the COM particle assumed here by *P0.0* is within the sustention polygon. *P0.0* represents the COM of the robot mass, it is a virtual particle and has no search swarm attached to it.

The swarms are conceived to work as follows: The memory particles are connected hierarchically from bottom to top, such a connection means that particle *(p0.2)*, *particle 2* of *swarm 0* communicates its position to both particles *(p0.1)* and *(p0.3)*, it gathers also the respective positions of these particles. Particle *(p0.3)* has only the position of *(p0.2)*. *P0* is a specific particle used to represent the body center of mass, *COM,* it's position is estimated using the best sub-swarms particles and using Eq(9)and (8). This particle is critic since the analysis of its coordinates will help to decide on whether the generated joints positions correspond to stable posture or not.

### 4.3 Stability control policy

Dynamic walking is the natural human attitude; the double support phases are very limited we move always on a single support. Static equilibrium walking can be described as a succession of standing stable positions, even in a simple support phase, if the walking is halted, during the cycle, the robot still standing up in an equilibrium position, Figure 5 illustrate the position of the COM during a walking cycle where simple support phases alternate with double support ones. For a humanoid it is easier to achieve a static equilibrium rather than a dynamic one. In a slow forward the static stability is assured if:

1) In a double support phase, the center of mass of the robot is placed in the center of the sustention polygon.

2) In a simple support phase, the center of mass COM is placed on the support foot. That condition insures that the robot will not fall down while it balances.

```
Initialize number of sub-Swarms.
Initialize number of particles of the sub-
swarms.
Initialize n1, n2 = max iteration number1.
Initialize the joints memories M(i,j).
Begin
   Loop
Select the support leg
Initialize the oscillating leg sub-swarms:
       Sub-swarm(i)= {Si.0, Si.1, Si.2}
Initialize the local search swarms S(i,j)
Loop
lunch Sub-swarms(i)
        for i= 1 to number of sub-Swrams
        run sub-swarms()

     evaluate local bests : Gbest(i,j)
       Next (i)
Validate the step:
       Estimate COM(x,y)position
       Check the stability of the walker
M(i,j) = Gbest (s(i,j).
Change support leg
Transfer the particles dynamics ()
   Until (number of iteration = n1)
   Until (number of iteration = n2)
   End
                (a)
```

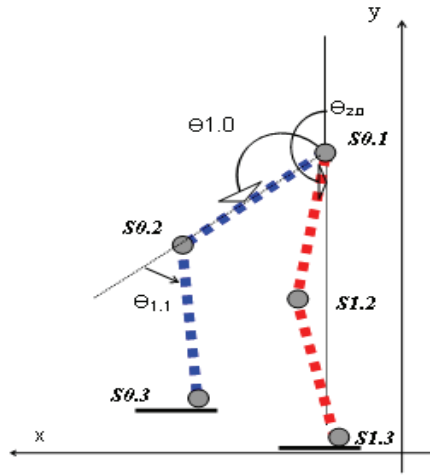

(c)

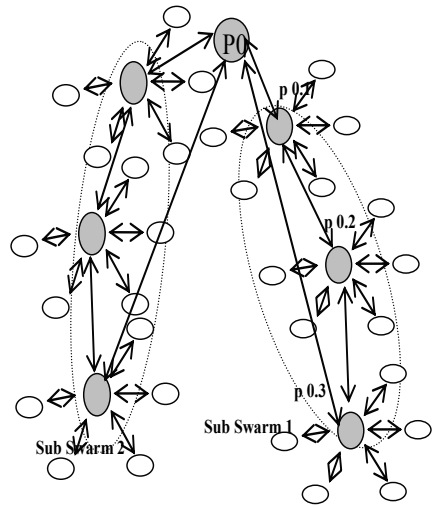

(b)

Fig. 4. (a) Pseudo-code of the proposed algorithm, (b) Sub swarms architecture and inference graph, the memories particles appears in grey, the search particles are white, (c) Side view of the simplified bipedal locomotion system.

From a biomechanics' point of view, the coordinates of the center of mass, COM, are estimated by Eq (9). If we consider a body with n segments ($S_1, S_2, \ldots, S_n$) and if we assume that the weight of these segments are ($m_1, m_2, \ldots m_n$) the center of mass can be estimated. In our case we need to estimate the coordinates of the CoM projection on the floor plan, which means that we need a pair of coordinates COM(x,y). The relative masses of the segments depend on the weight of the skeleton, on the nature and proportion of the used effectors and sensors; alternatively it can be directly inspired from human anthropomorphic studies even if human muscles still better than any industrial actuators.

$$x^i_{com} = \frac{\sum_{j=1}^{n\_Sw} m_j * x_j}{M} \quad (8)$$

$$y^i_{com} = \frac{\sum_{j=1}^{n\_Sw} m_j * y_j}{M} \quad (9)$$

Where :

*(i)* represents the iteration number,
*n_Sw* is the number of sub-Swarms in the proposed model, here *n_Sw* = 6, see figure 1(a). *M* is the mass of the locomotion system and $m_j$ represents the mass of the segment *($S_j$)*.

If we assume that the robot has a footprint which is propositional to its locomotion-system dimensions, and if we assume that the footprint is rectangular, see subsection A, we deduce a simple representation of the sustention polygon in both double and single support phases, see Figure 5. The foot-print is supposed to be rectangular with the length *fl* and a breadth *fb*, Eq(1) and (2). If only single support phases are used during the walking cycle the sustention polygon is limited to the segment joining *p0.3* (left ankle) to *p1.3* (right ankle), this is a constrained solution compared to the first one.

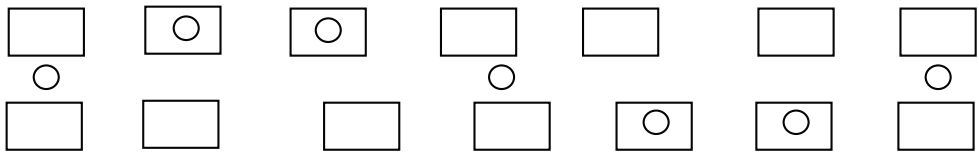

Fig. 5. A static walking cycle foot prints, COM (circle) projection on the footprints (rectangle).

### 4.4 Fitness functions

In our case we need both, local finesses functions in order to select local bests particles,

and a global fitness function that allow us to select the best posture within those assuming stability. Local best particles are those minimizing the error expressed by Eq (10).

$$f_{sw(j)} = \left\| \overrightarrow{p^{s}_{i,j} p^{th}_{i,j}} \right\| \qquad (10)$$

Where $f_{sw(j)}$ is the fitness function of the swarm (j), $p^{s}_{i,j}$ is the 3D point representing the memory particle of the joint (i,j); $p^{th}_{i,j}$ is the theoretical point corresponding to the joint position obtained by a direct kinematic solver.

$$p^{th}_{i,j} = \begin{bmatrix} x_{i-1} \\ y_{i-1} \\ z_{i-1} \end{bmatrix} + M^{sg}_{i} * M^{ft}_{i} * \begin{bmatrix} 0 \\ 0 \\ l \end{bmatrix} \qquad (11)$$

Where $M^{sg}_{(i-1)i}$ and $M^{sg}_{(i-1)i}$ represent respectively the rotation matrices on the lateral and frontal plans they are expressed as fellows:

$$M^{sg}_{i} = \begin{bmatrix} \cos(\theta_{j,i-1},) & 0 & -\sin(\theta_{j,i-1}) \\ 0 & 1 & 0 \\ \sin(\theta_{j,i-1}) & 0 & \cos(\theta_{j,i-1}) \end{bmatrix} \qquad (12)$$

Note that $\theta_{j,i-1}$ is the last stable position angle of the joint(i) that belongs to the sub-swarm(j) on lateral plan. The frontal plane rotation matrix for axis (Y,Z) is expressed by equation (13):

$$M^{ft}_{i} = \begin{bmatrix} 1 & 0 & 0 \\ 0 & \cos(\alpha_{j,i-1}) & -\sin(\alpha_{j,i-1}) \\ 0 & \sin(\alpha_{j,i-1}) & \cos(\alpha_{j,i-1}) \end{bmatrix} \qquad (13)$$

Where $\alpha_{j,i-1}$ : is the last stable position angle of the joint (i) that belongs to the sub-swarm(j) on the frontal plan.

To evaluate global best positions and stability policy we try to minimize the following expression, the best pattern is that one who minimizes the distance between the projection of the COM and the Gravity center on the sustention polygon.

$$f_{itness} = \sqrt{(y_{com} - y_{polycenter})^2} + \sqrt{(x_{com} - x_{polycenter})^2})  \qquad (14)$$

Note that $(x_{polycenter}, y_{polycenter})$ represent the coordinates of the sustention polygon center of gravity, they vary according to whether the robot is in single support or double support phase.

## 5. Discussions and Further Developments

In this paper we have briefly resumed main biped and humanoid locomotion research issues before introducing the IZIMAN project. We presented our experiments in human gaits captures, these gaits help us understanding the walking mechanism and are used as a comparative frame work to validate the simulated approaches; classical kinematics modeling is also used to generate joints trajectories but is not detailed in this paper. We essentially detailed our biologically inspired hybrid gaits generation methodology. The proposal is based on particle swarm optimization.
The joints extracted from the biomechanics experimentations are limited since only six walkers were involved in the experimentation process, in the future large scale gaits captures will be organized; even if this work had been done by earlier bipedal locomotion researches it still very instructive on the way walking and anthropomorphisms works.
Particle swarm optimization belongs to what is commonly assumed to be evolutionary computing; it is based on a quiet simple equation fast to compute with a low memory cost, they can be used as an alternative to mathematical equation solvers especially in non linear systems.
3D walking steps are shown in figure6 (a), while a lateral walking is represented in figure6 (b). The walking gaits of the COM can be observed in figure 6(c), evolutionary based ones appear in figure6 (d). The PSO based gait generator performs globally similar results to those obtained by classical modeling (Ammar, 2006) on the axes x and y, while the z-gaits are slightly different showing that the walker has a dissymmetric motion; one of the members oscillates longer than the other. In human normal walking attitude, a slight dissymmetry is also observed but it is not so important than that produced by the PSO approach. On the other hand and assuming that PSO is a non deterministic technique, only valid gaits should be saved in memory. The convergence of the algorithm does not insure that a complete walking cycle is always gathered..! A learning technique should be soon introduced to overcome this problem. The obtained gaits will be soon implemented as reference joint trajectories in a small size humanoid assembled in the REGIM laboratory using the BIOLOID expert robotics kit.

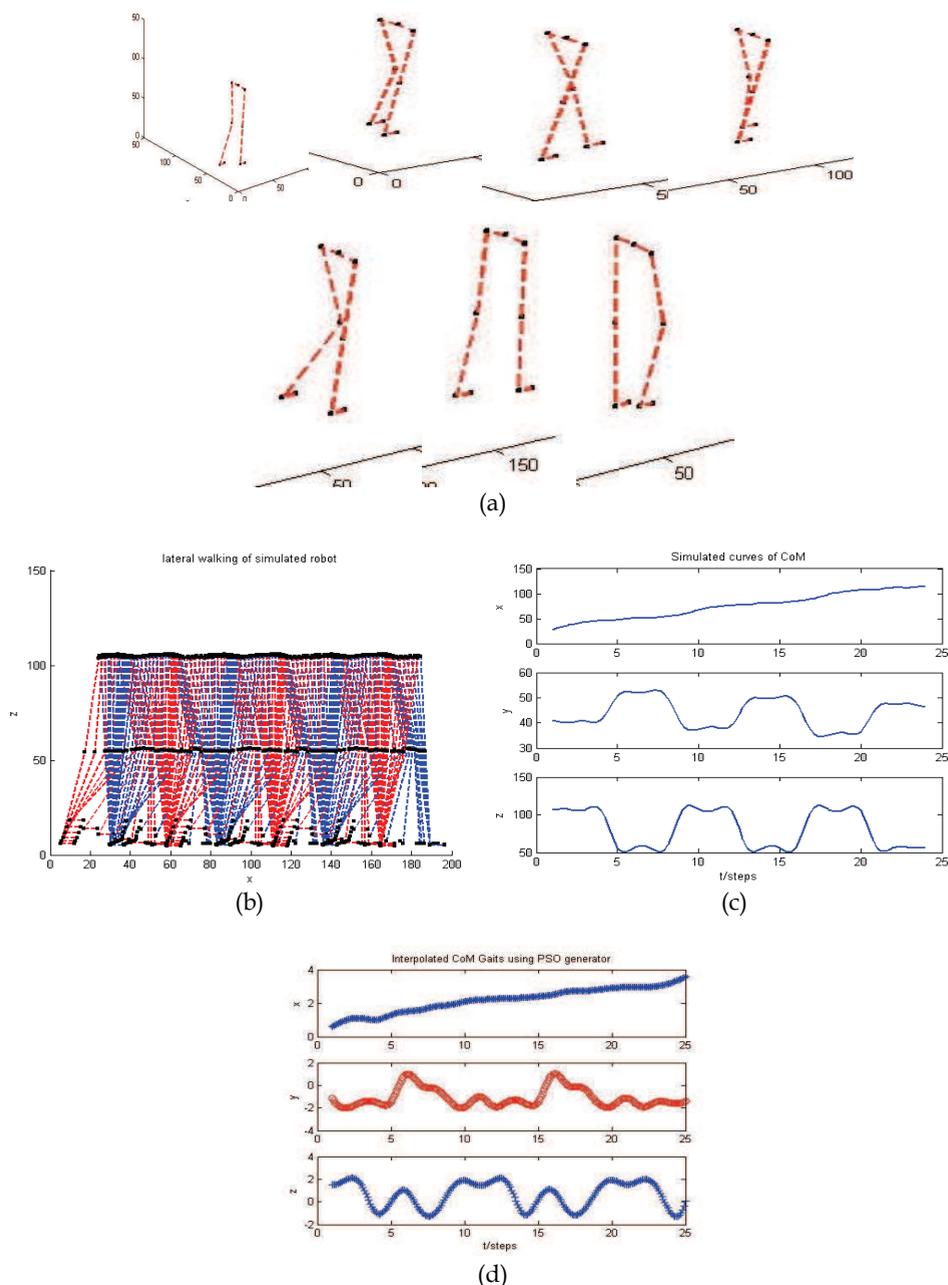

Fig. 6. Simulation results, (a) walking frames screen shots, (b) lateral gaits of segments femur and tibia from classical kinematics' simulation, (c) COM gaits from classical kinematics' simulation, (d) COM gaits from PSO proposal approach.

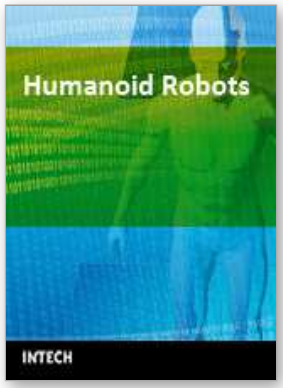

**Humanoid Robots**
Edited by Ben Choi

ISBN 978-953-7619-44-2
Hard cover, 388 pages
**Publisher** InTech
**Published online** 01, January, 2009
**Published in print edition** January, 2009

Humanoid robots are developed to use the infrastructures designed for humans, to ease the interactions with humans, and to help the integrations into human societies. The developments of humanoid robots proceed from building individual robots to establishing societies of robots working alongside with humans. This book addresses the problems of constructing a humanoid body and mind from generating walk patterns and balance maintenance to encoding and specifying humanoid motions and the control of eye and head movements for focusing attention on moving objects. It provides methods for learning motor skills and for language acquisition and describes how to generate facial movements for expressing various emotions and provides methods for decision making and planning. This book discusses the leading researches and challenges in building humanoid robots in order to prepare for the near future when human societies will be advanced by using humanoid robots.

**How to reference**
In order to correctly reference this scholarly work, feel free to copy and paste the following:

Nizar Rokbani, Boudour Ammar Cherif and Adel M. Alimi (2009). Toward Intelligent Biped-Humanoids Gaits Generation, Humanoid Robots, Ben Choi (Ed.), ISBN: 978-953-7619-44-2, InTech, Available from: http://www.intechopen.com/books/humanoid_robots/toward_intelligent_biped-humanoids_gaits_generation

# INTECH
open science | open minds